\documentclass[runningheads]{llncs}
\makeatletter
\newcommand{\printfnsymbol}[1]{%
	\textsuperscript{\@fnsymbol{#1}}%
}
\usepackage{bbding}
\usepackage{graphicx}
\usepackage{amssymb}
\usepackage{amsmath}
\usepackage{cite}
\usepackage{CJKutf8}
\usepackage{multirow}
\usepackage{url}
\usepackage{xcolor}

\makeatletter
\def\@fnsymbol#1{\ensuremath{\ifcase#1\or *\or \dagger\or \ddagger\or
		\mathsection\or \mathparagraph\or \|\or **\or \dagger\dagger
		\or \ddagger\ddagger \else\@ctrerr\fi}}
\makeatother

\begin{document}

	\title{Overview of the NLPCC 2025 Shared Task 4: Multi-modal, Multilingual, and Multi-hop Medical Instructional Video Question \\ Answering Challenge}
	%
	%
	
	\vspace{-1.5cm}
		\author{Bin Li\inst{1}\thanks{These authors contribute this work equally.} \and
			Shenxi Liu \inst{2}\printfnsymbol{1} \and 
			Yixuan Weng\inst{3} \and 
			Yue Du\inst{1} \and \\
			Yuhang Tian\inst{2} \and 
			Shoujun Zhou \inst{1}\thanks{Corresponding author; Email: sj.zhou@siat.ac.cn.}
}

	 	%
		\institute{Shenzhen Institute of Advanced Technology, Chinese Academy of Sciences, \\
				\email{\{b.li2, yue.du2, sj.zhou\}@siat.ac.cn,} \\			\and
		School of Computer Science and Technology, Beijing Institute of Technology, 		\\
				\email{\{liushenxi, tianyuhang\}@bit.edu.cn,} \\	\and
			School of Engineering, Westlake University \\
				\email{wengsyx@gmail.com} \\
}	 	
		\authorrunning{Li et al.}

\titlerunning{Overview of the NLPCC 2025 Shared Task 4}
%
\maketitle              
\vspace{-0.5cm}
\begin{abstract}
	Following the successful hosts of the 1-st (NLPCC 2023 Foshan) CMIVQA and the 2-rd (NLPCC 2024 Hangzhou) MMIVQA challenges, this year, a new task has been introduced to further advance research in multi-modal, multilingual, and multi-hop medical instructional question answering (M4IVQA) systems, with a specific focus on medical instructional videos. The M4IVQA challenge focuses on evaluating models that integrate information from medical instructional videos, understand multiple languages, and answer multi-hop questions requiring reasoning over various modalities. This task consists of three tracks: multi-modal, multilingual, and multi-hop Temporal Answer Grounding in Single Video (M4TAGSV), multi-modal, multilingual, and multi-hop Video Corpus Retrieval (M4VCR) and multi-modal, multilingual, and multi-hop Temporal Answer Grounding in Video Corpus (M4TAGVC). Participants in M4IVQA are expected to develop algorithms capable of processing both video and text data, understanding multilingual queries, and providing relevant answers to multi-hop medical questions. We believe the newly introduced M4IVQA challenge will drive innovations in multimodal reasoning systems for healthcare scenarios, ultimately contributing to smarter emergency response systems and more effective medical education platforms in multilingual communities\footnote[1]{Official Website: \url{https://cmivqa.github.io/}.}.
\end{abstract}
\section{Introduction}
Recent advancements in AI-assisted healthcare have demonstrated potential across various clinical scenarios \cite{li2023overview, li2024overview,wang2025systematic}, particularly in imaging diagnosis \cite{lu2023tf,yu2025prnet} and personalized treatment consultation \cite{yu2024scnet,bavcic2024towards}. The development of Video Question Answering (VideoQA) technology in the medical domain has achieved significant breakthroughs in extracting information from video content and answering related queries \cite{wen2024learning,bi2025prismselfpruningintrinsicselection}. However, significant gaps persist in the development of reasoning systems capable of handling multimodal domain knowledge \cite{lu2024robust,li2025set}. Existing medical VideoQA systems still face three fundamental challenges: (1) temporal grounding of cross-modal evidence, (2) multilingual semantic alignment, and (3) multi-hop reasoning across the video corpus.
\par
To advance research and application in this field, NLPCC 2025 has introduced the Multi-modal, Multilingual, and Multi-hop Medical Instructional Question Answering (M4IVQA) challenge, specifically focusing on instructional medical videos. This task includes three tracks:: multi-modal, multilingual, and multi-hop Temporal Answer Grounding in Single Video (M4TAGSV), multi-modal, multilingual, and multi-hop Video Corpus Retrieval (M4VCR) and multi-modal, multilingual, and multi-hop Temporal Answer Grounding in Video Corpus (M4TAGVC). The main goals of the M4IVQA challenge include:
\begin{enumerate}
	\item \textbf{Multi-hop reasoning}: The task requires systems to perform multi-step inference across heterogeneous information sources within or across videos. This reflects the complexity of real medical questions, which often require combining multiple pieces of evidence rather than relying on a single visual cue \cite{xia2022lingyi}.
	
	\item \textbf{Multimodal fusion}: Effective performance depends on the model’s ability to jointly understand and integrate diverse modalities \cite{bi2025llavasteeringvisualinstruction}, including instructional narration \cite{weng2023visual}, visual demonstrations \cite{zhong2025enhancing}, and on-screen text \cite{qiu2025generative} (e.g., captions or annotations), which are all essential components in medical video content.
	
	\item \textbf{Cross-video temporal grounding}: This involves not only retrieving the most semantically relevant videos, but also accurately pinpointing the temporal boundaries of answer-containing segments. Such a capability is critical in medical education and practice, where relevant knowledge may be dispersed across different procedural videos or case demonstrations.
\end{enumerate}
\par
The design of the M4IVQA task is closely aligned with the practical demands of medical education and emergency information access. In surgical 
scenarios, the ability to rapidly locate key operational procedures within videos can be life-saving. In educational contexts, intelligent question-answering systems empower learners to better comprehend complex medical techniques by referencing precise visual evidence \cite{jing2023multimodal,ke2025detection}. Moreover, multilingual support ensures that high-quality medical instructional content can transcend language barriers, thereby reaching and benefiting a broader global audience.
\par
This paper provides a comprehensive overview of the M4IVQA task, including its design motivations, dataset composition, evaluation framework, and potential application scenarios. Through this challenge, we aim to advance research in medical AI, multimodal reasoning, and cross-lingual understanding, ultimately contributing to improved global medical training and more accessible dissemination of emergency knowledge.
\section{Task Introduction}
\vspace{-0.1cm}
\subsection{Definition of Each Track}
The M4IVQA challenge aims to promote the development of medical instructional video question-answering technology, especially in reasoning ability. This task requires participants to develop algorithms capable of processing both video and text data, understanding user's queries \cite{he2025enhancing}, and providing relevant answers to multi-hop medical questions. The M4IVQA task is divided into three challenging subtasks, which is described as follows.
\begin{enumerate}
	\begin{figure*}[!h]
		\centering
				\vspace{-0.4cm}
		\includegraphics[scale=0.47]{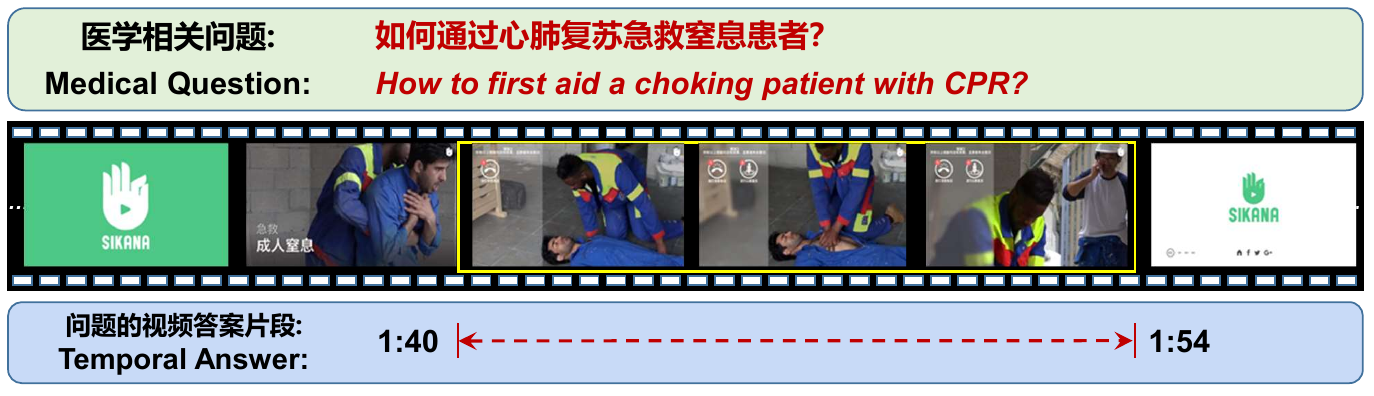}
		\caption{Introduction of the multi-modal, multilingual, and multi-hop Temporal Answer Grounding in Singe Video (M4TAGSV) track.} 
		\label{taskdes1}
		\vspace{-0.6cm}
	\end{figure*}
	
		\begin{figure*}[!h]
		\centering
		\vspace{-0.5cm}
		\includegraphics[scale=0.47]{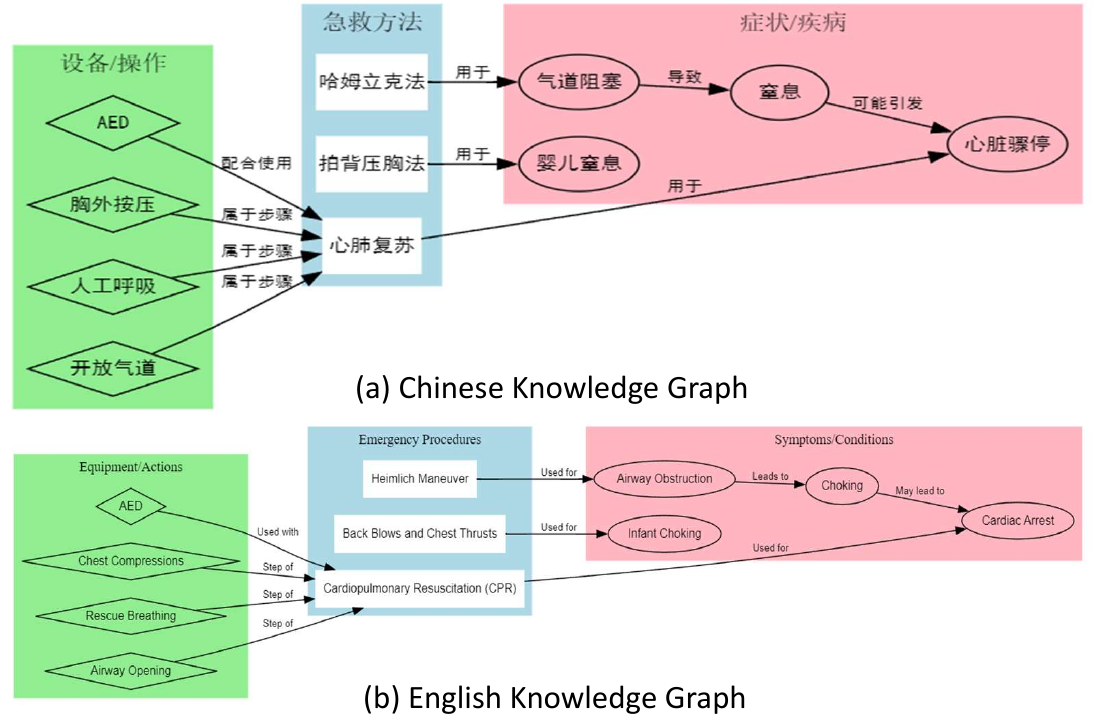}
		\caption{Illustration of the knowledge graph derived from the given medical video.} 
		\label{taskdes1111}
		\vspace{-0.8cm}
	\end{figure*}
	\item M4TAGSV: This track targets temporal answer grounding in a single video. Given a natural‑language question in both Chinese or English language, AI systems must precisely identify the contiguous temporal span that contains the requisite visual evidence. Accordingly, the track benchmarks fine‑grained video comprehension and high‑resolution temporal localization, thereby laying the groundwork for the more complex corpus‑level tasks. 
	As illustrated in Figure~\ref{taskdes1}, for the Chinese query “{\begin{CJK}{UTF8}{gbsn}如何通过心肺复苏急救窒息患者？\end{CJK}}” or its English counterpart “How to first aid a choking patient with CPR?”, the correct prediction is the segment {01:40-01:54}. We also prepared knowledge graphs in different languages, which are derived from Wikipedia or common knowledge bases \cite{lu2024mace}. As shown in Fig. \ref{taskdes1111}, the ``CPR" in the knowledge graph helps the model understand professional medical terms and perform more complex multi-hop reasoning.

	\begin{figure*}[t]
	\centering
			\vspace{-0.1cm}
	\includegraphics[scale=0.44]{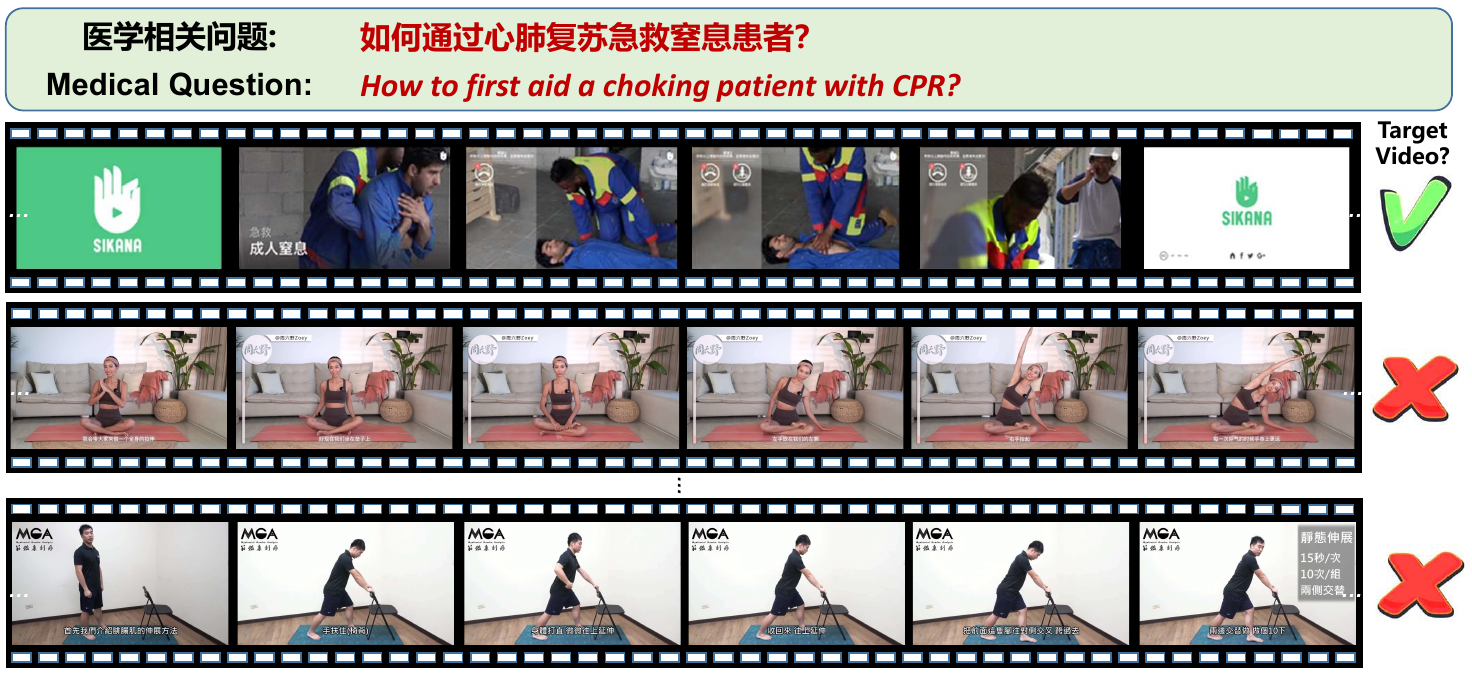}
			\vspace{-0.4cm}
	\caption{Introduction of the multi-modal, multilingual, and multi-hop Video Corpus Retrieval (M4VCR) track.} 
					\vspace{-0.55cm}
	\label{taskdes2}
\end{figure*}
	\item M4VCR: This track integrates a domain-specific knowledge graph to facilitate structured video retrieval, demanding not only robust cross-lingual comprehension but also efficient processing of large-scale multimodal data encompassing visual, textual, and linguistic information. Performance is evaluated by assessing the relevance ranking of returned videos, measuring the system’s capacity for information retrieval and semantic relevance judgment across heterogeneous sources.
	As illustrated in Fig. \ref{taskdes2}, consider a Chinese query: “{\begin{CJK}{UTF8}{gbsn}如何通过心肺复苏急救窒息患者？\end{CJK}}” or an English query: “How to first aid a choking patient with CPR?”. The task requires identifying the most semantically aligned video within the corpus. 
	\begin{figure*}[!h]
	\centering
			\vspace{-0.1cm}
	\includegraphics[scale=0.47]{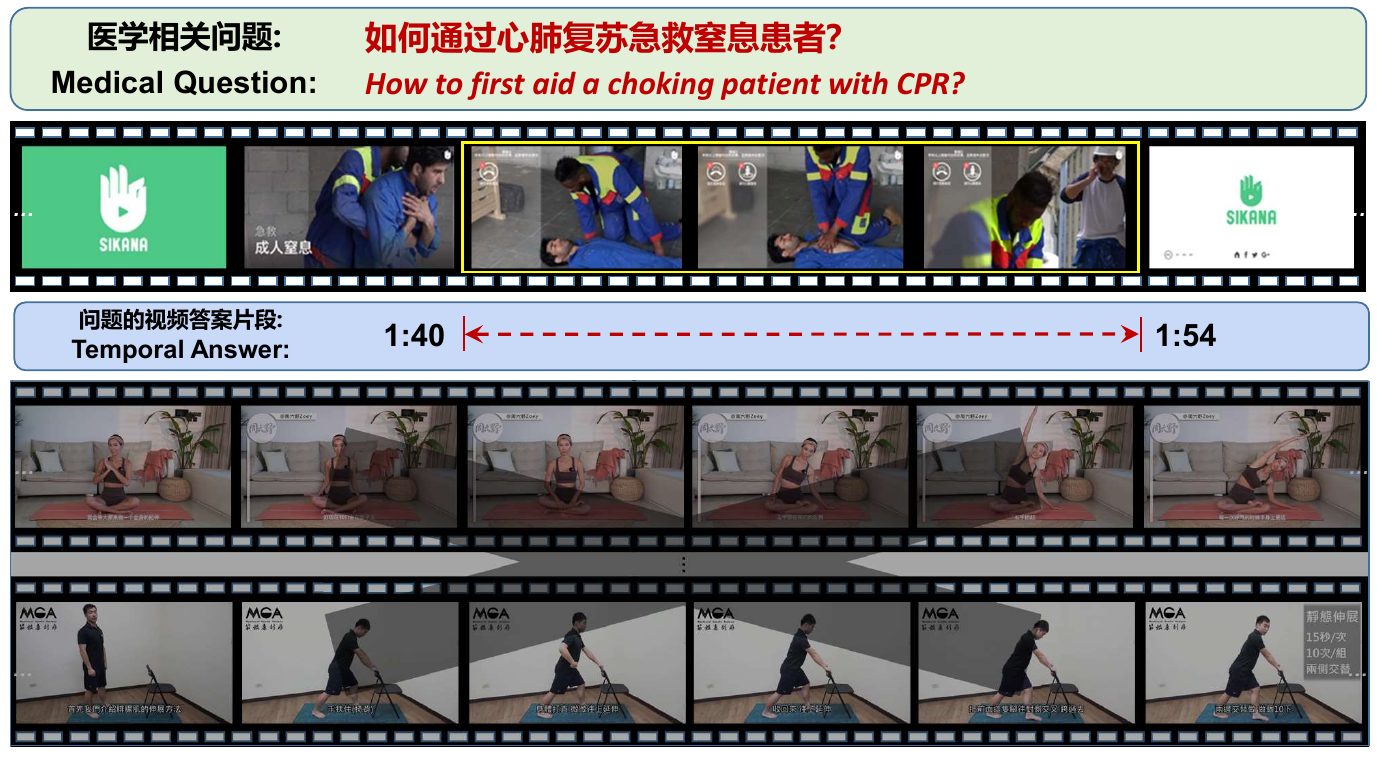}
				\vspace{-0.3cm}
	\caption{Introduction of the multi-modal, multilingual, and multi-hop Temporal Answer Grounding in Video Corpus (M4TAGVC) track.} 
	\label{taskdes3}
			\vspace{-0.5cm}
\end{figure*}

	\item M4TAGVC: This task combines the key challenges of the first two tracks, requiring systems to retrieve relevant videos from large-scale corpora and precisely localize answer-containing temporal segments (e.g., ``visual answers''). This task assesses holistic capabilities in video comprehension, cross-corpus information retrieval, cross-lingual semantic alignment, and temporal grounding—integrating multi-hop reasoning across visual, textual, and linguistic data. As shown in Fig. \ref{taskdes3}, given a query like the Chinese “{\begin{CJK}{UTF8}{gbsn}如何通过心肺复苏急救窒息患者？\end{CJK}}” or the English ``How to first aid a choking patient with CPR?'', the system shall first identify semantically matched videos and then pinpoint the exact time span (e.g., 1:40–1:54s) where the answer is demonstrated. This requires a model for corpus-level retrieval and fine-grained temporal localization.
\end{enumerate}

\subsection{Evaluation Metrics}
The M4IVQA challenge utilizes a holistic framework of evaluation metrics that assess participating systems through multiple dimensions. These metrics not only examine the correctness of system outputs but also quantify their precision levels and ranking capabilities, ensuring a thorough and equitable assessment process.
\begin{enumerate}
\item M4TAGSV: Our primary evaluation metric is the Intersection over Union (IoU), which measures the overlap extent between predicted and ground-truth regions of interest \cite{weng2023visual}. To achieve a holistic performance assessment across all test scenarios, we compute the mean IoU (mIoU). Drawing from prior research, we integrate supplementary evaluation metrics. A key addition is the ``R@n, IoU = 
$\mu$" metric, which evaluates instances where the overlap between predicted temporal spans and ground-truth moments surpasses a threshold $\mu$ among the top-n retrieved segments. In our specific evaluation framework, we fix n at 1 and examine $\mu$ values of 0.3, 0.5, and 0.7. These metrics are formalized by the following equations.
\begin{equation}
	\begin{aligned}
	\mathrm{IOU} & =\frac{A \cap B}{A \cup B}  \\
		\operatorname{mIOU} &= \left(\sum_{i=1}^N \mathrm{IOU}_i\right) / N
	\end{aligned}
	\label{e1}
\end{equation}
where A and B represent different spans, and $\sum_{i=1}^N \mathrm{IOU}_i$ represent IoU = 0.3/0.5/0.7 respectively, $N$=3. 
\item M4VCR: We utilized the standard evaluation metrics in video retrieval \cite{yi2025score,10096391}, specifically the ``R@n" metric with n set to 1, 10, and 50. This metric quantifies the system's recall performance at varying retrieval depths. To assess the effectiveness of multilingual medical instructional video retrieval, we incorporated the Mean Reciprocal Rank (MRR) score. As a holistic measure, MRR evaluates both the correctness of retrieved results and the positional ranking of accurate entries, defined by the following calculation:

\begin{equation}
	MRR=\frac{1}{|V|} \sum_{i=1}^{|V|} \frac{1}{\operatorname{Rank}_i}
\end{equation}

Here, $|V|$ represents the number of videos in the corpus. For each testing sample $V_i$, Rank$_i$ denotes the position of the ground-truth target video in the predicted list.

In this track, the primary ranking metric is the ``verall" score, which combines the R@1, R@10, R@50, and MRR scores:

\begin{equation}
	\text { Overall }=\frac{1}{|M|} \sum_{i=1}^{|M|} {\text { Value}_i}
\end{equation}

Here, $|M|$ is the number of evaluation metrics (4 in this case, corresponding to R@1, R@10, R@50, and MRR), and Value$_i$ denotes the $i$-th metric score.
\item M4TAGVC: We retained the Intersection over Union (IoU) metric in the first task to evaluate the precision of temporal localization. Additionally, the second task incorporates retrieval-oriented metrics, including the ``R@n" (with n set to 1, 10, and 50) and mean reciprocal rank (MRR) scores. To integrate these, we introduce the ``R@1|mIoU, R@10|mIoU, and R@50|mIoU" metrics, which compute the average IoU values at different retrieval depths, offering a comprehensive assessment of model performance. Participating models’ final rankings in this task are primarily determined by a composite score termed ``Average," derived from the mean of three indicators (denoted as $\text { Value}^{\prime}_i$ for the $i$-th indicator): R@1|mIoU, R@10|mIoU, and R@50|mIoU. The specific calculation formula is as follows:

\begin{equation}
	\text { Average }=\frac{1}{\left|M^{\prime}\right|}  \sum_{i=1}^{\left|M^{\prime}\right|} {\text { Value}^{\prime}_i}
\end{equation}
where the $|M^{\prime}|$ = 3.
\end{enumerate}

\subsection{Dateset}
	
	\begin{figure*}[ht]
	\centering
	\includegraphics[scale=0.83]{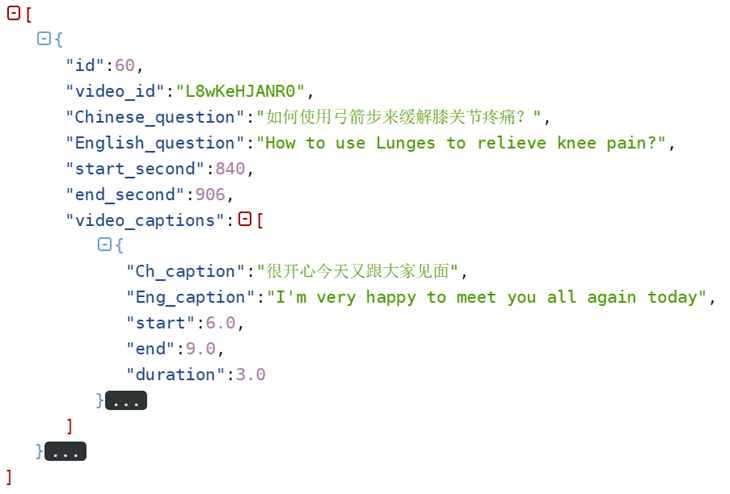}
				\vspace{-0.5cm}
	\caption{Dataset examples of the M4IVQA shared task.} 
	\label{taskdes11}
			\vspace{-0.6cm}
\end{figure*}
The M4IVQA challenge builds upon the dataset of MMIVQA \cite{li2024overview} and incorporates new knowledge graphs, utilizing a specially constructed multilingual medical instructional video dataset with unique and comprehensive characteristics, as illustrated in Fig. \ref{taskdes11}. Sourced from medical educational channels on YouTube via the Pytube tool\footnote[2]{https://github.com/pytube/}, the dataset covers diverse topics such as first aid, medical emergency management, and medical education, ensuring professional and varied content. A core feature is its multilingual design: each video includes both Chinese and English subtitles directly extracted from YouTube, alongside corresponding questions and answers in both languages.
Chinese questions (Chinese\_question) are manually crafted by Chinese medical experts, while English questions (English\_question) are translated and reviewed by native English-speaking physicians to guarantee linguistic accuracy and professional rigor. The dataset’s foundation in MMIVQA, combined with the integration of new knowledge graphs, enriches its structure for comprehensive multimodal and multilingual evaluation.
\par
To ensure data quality, annotations are meticulously conducted by professionals with medical backgrounds. Each video may contain multiple question-answer pairs, where semantically equivalent questions map to a single unique answer. Every question is annotated with precise answers and timestamps (start and end seconds) to anchor responses to specific video segments. Additionally, the dataset includes automatically generated Chinese and English subtitles via Whisper\footnote[3]{https://github.com/openai/whisper}, providing participants with supplementary textual context beyond the video content.  
A distinctive feature of the dataset is its detailed sample annotations for the M4TAGV task, including ``id" (for video retrieval indexing) and YouTube’s unique ``video\_id". The ``id" identifiers remain non-public during the competition to maintain evaluation fairness. For comprehensive dataset details and download links, participants can access the GitHub repository at \url{https://github.com/cmivqa/NLPCC2025_M4IVQA}, which offers extensive resource support for model development and testing. \par \par
 \begin{table}[ht!]
	 	\label{datasets11}
	 		 	\caption{Details of the datasets in NLPCC shared task 4.}
	 	\centering \small
	 	\renewcommand\arraystretch{1.2}	\setlength{\tabcolsep}{0.3mm}
	\begin{tabular}{ccccccc}
		\noalign{\hrule height 1pt}
		Dataset                             & Videos & QA pairs & KG Nums. & Ch\_Q.AvgLen. & Eng\_Q.AvgLen. & Video Avg. Len. \\
		\noalign{\hrule height 0.5pt}
		Train                       & 1228   & 5840     & 28490       & 17.16           & 6.97  & 263.3           \\
		Dev & 200    & 983      & 4637       & 17.81             & 7.26  & 242.4           \\
		Test  & 200    & 1022      & 5011       & 17.48           &  7.44 & 310.9          \\
		\noalign{\hrule height 1pt}
	\end{tabular}
			\vspace{-0.5cm}
	 	\label{t1}
\end{table}
\par
All the Train, Dev and Test files in the M4IVQA dataset include videos, audio, and corresponding subtitles. To maintain consistency, all the Chinese questions and subtitles have been converted to Simplified Chinese.
As shown in Table 1, the dataset is divided into three subsets: Train, Dev, and Test sets. The Train set contains 1228 videos with 5840 QA pairs, the Dev set includes 200 videos with 983 QA pairs, and the Test set consists of 200 videos with 1022 QA pairs. The Knowledge Graph (KG) number (Triple) for train set is 28490, 4637 for the Dev set, and 5011 for the Test set. The average length of Chinese questions (AvgLen. Ch\_Q) is 17.16 characters for the Train set, 17.81 for the Dev set, and 17.48 for the Test set. For English questions (Eng\_Q AvgLen.), the average lengths are 6.97, 7.26, and 7.44 words for Train, Dev, and Test sets respectively. The average video length (Video Avg. Len.) varies across the sets, with 263.3 seconds for Train set, 242.4 seconds for Dev  set, and 310.9 seconds for Test  set.
			\vspace{-0.2cm}
\section{Evaluation Results}
There are a total of 21 teams registered for the NLPCC 2025 Shared Task 4. During the testing phase, 10 teams submitted their results. Here is a brief introduction to the representative systems designed by \textbf{Baichuan} for track 1, \textbf{DIMA} for track 2, and \textbf{MedEcho} for track 3.
\begin{table}[!ht]
				\vspace{-0.4cm}
	 	\caption{Final Results of the track 1.}
		 	\centering \small
	\renewcommand\arraystretch{1.2}	\setlength{\tabcolsep}{0.30mm}
	\begin{tabular}{l|c|c|c|c|c}
		\noalign{\hrule height 1pt}
		Rank & Team ID & R@1,IoU=0.3 & R@1,IoU=0.5 & R@1,IoU=0.7 & mIoU(R@1) 	
		\\	\noalign{\hrule height 0.5pt}
		 1  & Baichuan & 0.5133&	0.3612	&0.2103	&0.3717 \\ 
		 2  & Ouc\_AI \cite{zhang2024improving} & 0.5088&	0.3542	&0.2054	&0.3637 \\ 
		 3  & SEU & 0.4719& 0.3314	&0.1973	&0.3411 \\ 
		4   & SETAG\cite{zhou2023improving} & 0.4775&	0.3209&	0.1898&	0.3389 \\ 
		5 & BIGG  & 0.4099&	0.2863	&0.1544	&0.2977 \\ 
		6  & PIPINSTALL \cite{li2024overview} & 0.4055&	0.2911&	0.1454&	0.2898 \\ 
		7 & Random Pick Method \cite{li2023overview}  & 0.0571&		0.0465	&	0.0358	&	0.0397 \\ 
	\noalign{\hrule height 1pt}
	\end{tabular}
			\vspace{-0.5cm}
\end{table}

\begin{table}[!ht]
				\vspace{-0.4cm}
		 	\caption{Final Results of the track 2.}
		 	\centering \small
\renewcommand\arraystretch{1.2}	\setlength{\tabcolsep}{1.85mm}
	\begin{tabular}{l|c|c|c|c|c|c}
\noalign{\hrule height 1pt}
		Rank & Team ID & R@1 & R@10 & R@50 & MRR & Overall \\ \noalign{\hrule height 0.5pt}
		1  &DIMA& 0.3264&	0.4211&	0.5177&	0.3407&	1.6059 \\ 
		2  &NYU& 0.3213&	0.4137&	0.5104&	0.3354&	1.5808 \\ 
		3  &sun \cite{yu2024mqua}& 0.3121&	0.4078&	0.4966&	0.3245&	1.5410 \\ 
		4  &DSG-1 \cite{lei2023two}& 0.2644&	0.3545&	0.4414&	0.2887&	1.3491 \\  
		5  &Wjh& 0.2744&	0.3312&	0.4117&	0.2551&	1.2724 \\
		6  &GEN \cite{li2024overview}& 0.1311&	0.1074&	0.0978&	0.1142&	0.4505 \\ 
		7  & Random Pick Method \cite{li2023overview} & 0.03438&	0.03668&	0.05238&	0.04428&	0.1674 \\ 

	\noalign{\hrule height 1pt}
	\end{tabular}

\end{table}

\begin{table}[!ht]
				\vspace{-0.2cm}
		\caption{Final Results of the track 3.}
						\vspace{-0.1cm}
	\centering \small
	\renewcommand\arraystretch{1.2}	\setlength{\tabcolsep}{1.5mm}
    \begin{tabular}{l|c|c|c|c|c}
\noalign{\hrule height 1pt}
	Rank & Team ID & R@1$|$mIoU & R@10$|$mIoU & R@50$|$mIoU & Average \\ \noalign{\hrule height 0.5pt}
	1 & MedEcho& 0.1284& 	0.2417& 	0.3243& 	0.2314 \\ 
	2 & NYU& 0.1202& 	0.2366& 	0.3133& 	0.2233 \\ 
	3 & IIEleven \cite{ma2024multilingual}& 0.1141& 	0.2343& 	0.3098& 	0.2200 \\ 
	4 & Nsddd \cite{cheng2023unified}& 0.1078& 	0.2145& 	0.2997& 	0.2112 \\ 
	5 & Random Pick Method \cite{li2023overview}& 0.0247& 	0.0397& 	0.0509& 	0.0384 \\ 

\noalign{\hrule height 1pt}
	\end{tabular}
			\vspace{-0.5cm}
\end{table}



For Track 1, the Baichuan team developed a RAG - enhanced multi - modal fusion framework. They take medical videos, queries, subtitles, and knowledge graphs as inputs, where RAG enriches the query and subtitle text by retrieving relevant information from the knowledge graph. The CLIP aligns video frames and queries to get a rough time interval. Then, we adopt the feature extractor to separately extract visual and textual features \cite{cui2024correlation}. Pseudo - label generators guide the training of visual and textual predictors. A mutual knowledge transfer module is used between modalities to improve the prediction accuracy of start and end times. The whole framework is optimized through a supervised learning loop to achieve accurate localization of answers in medical videos.
\par
For the track 2, the DIMA team proposed a three-stage retrieval-rerank framework. In the 1-st stage, subtitle chunks are enriched with knowledge-graph facts, encoded by LaBSE \cite{feng2020language}, and organized into a hierarchical index. In the 2-nd stage, a similarity-based traversal prunes low-score branches of this index and returns a compact pool of top candidate segments. In 3-rd stage, a lightweight multilingual LLM reranks these segments, and max-pool aggregation yields the final video ranking, augmenting the fine-grained multilingual relevance.
\par
For Track-3, the MedEcho team proposed a multi-hop knowledge-enhanced cross-modal retrieval framework designed for multilingual medical instructional video question answering. The approach begins by extracting visual and textual features using CLIP-ViT and a BERT-based encoder \cite{li2022vpai_lab}, respectively. Each video is decomposed into aligned frame-subtitle pairs, and a transformer-based cross-modal alignment module is applied to fuse these features and capture fine-grained semantic relationships \cite{jing2023category}. To address modality imbalance and enable efficient retrieval, a subtitle-level embedding library is constructed through temporal modeling. Moreover, the method retrieves relevant knowledge triples from an external medical knowledge graph and performs multi-hop reasoning via a RAG module to generate an enhanced query representation that better reflects the underlying medical intent. During training, contrastive learning is conducted between the enhanced query and both positive and negative subtitle segments to improve retrieval robustness. At inference time, they compute cosine similarity between the enhanced query vector and subtitle embeddings, with the final answer span determined by the position and value of the highest-scoring match in the similarity matrix.
\vspace{-0.4cm}
\section{Conclusion}
\vspace{-0.2cm}
This paper comprehensively overviews the NLPCC 2025 Shared Task 4: Multi-modal, Multilingual, and Multi-hop Medical Instructional Video Question Answering (M4IVQA). We introduced this innovative task, which comprises three challenging tracks: Multi-modal, Multilingual, Multi-hop Temporal Answer Gr-ounding in Single Video (M4TAGSV), Multi-modal, Multilingual, Multi-hop Video Corpus Retrieval (M4VCR), and Multi-modal, Multilingual, Multi-hop Temporal Answer Grounding in Video Corpus (M4TAGVC). These tracks were designed to push the boundaries of multi-modal fusion, multi-hop reasoning, and cross-lingual semantic alignment in the medical domain. We detailed the specially curated dataset used for this competition, highlighting its unique features such as bilingual (Chinese and English) content, professional medical annotations, cross-modal alignment, and diverse medical scenarios with multi-modal instructional elements. While the results of this competition demonstrate significant progress in multi-modal and cross-lingual medical video question answering, there remains substantial room for improvement before such systems can be deployed in real-world applications. The challenge of accurately integrating multi-modal evidence, performing multi-hop reasoning across video corpora, and achieving accurate visual answer localization, proves to be a complex task that requires further research and development. In conclusion, we believe that the M4IVQA task has opened up new avenues for research at the intersection of medical AI, multi-modal learning, cross-lingual understanding, and multi-hop reasoning, fostering advancements in global medical training and emergency knowledge dissemination.
			\vspace{-0.2cm}
\section{Acknowledge}
This work was supported by the Natural Science Foundation of Guangdong Province (No. 2023A1515010673), in part by the Shenzhen Science and Technology Innovation Bureau key project (No. JSGG20220831110400001, No. CJGJZD \\ 20230724093303007,KJZD20240903101259001), in part by Shenzhen Medical Research Fund (No. D2404001), in part by Shenzhen Engineering Laboratory for Diagnosis \& Treatment Key Technologies of Interventional Surgical Robots (XM \\ HT20220104009), and the Key Laboratory of Biomedical Imaging Science and System, CAS, for the Research platform support.
%
%
%
%
\vspace{-0.2cm}

\bibliographystyle{unsrt}
\bibliography{a}

\begin{thebibliography}{10}

\bibitem{li2023overview}
Bin Li, Yixuan Weng, Hu~Guo, Bin Sun, Shutao Li, Yuhao Luo, Mengyao Qi, Xufei
  Liu, Yuwei Han, Haiwen Liang, et~al.
\newblock Overview of the nlpcc 2023 shared task: Chinese medical instructional
  video question answering.
\newblock In {\em CCF International Conference on Natural Language Processing
  and Chinese Computing}, pages 233--242. Springer, 2023.

\bibitem{li2024overview}
Bin Li, Yixuan Weng, Qiya Song, Lianhui Liang, Xianwen Min, and Shoujun Zhou.
\newblock Overview of the nlpcc 2024 shared task 7: Multi-lingual medical
  instructional video question answering.
\newblock In {\em CCF International Conference on Natural Language Processing
  and Chinese Computing}, pages 429--439. Springer, 2024.

\bibitem{wang2025systematic}
Yiting Wang, Jiachen Zhong, and Rohan Kumar.
\newblock A systematic review of machine learning applications in infectious
  disease prediction, diagnosis, and outbreak forecasting.
\newblock 2025.

\bibitem{lu2023tf}
Shilin Lu, Yanzhu Liu, and Adams Wai-Kin Kong.
\newblock Tf-icon: Diffusion-based training-free cross-domain image
  composition.
\newblock In {\em Proceedings of the IEEE/CVF International Conference on
  Computer Vision}, pages 2294--2305, 2023.

\bibitem{yu2025prnet}
Xinlei Yu, Ahmed Elazab, Ruiquan Ge, Jichao Zhu, Lingyan Zhang, Gangyong Jia,
  Qing Wu, Xiang Wan, Lihua Li, and Changmiao Wang.
\newblock Ich-prnet: a cross-modal intracerebral haemorrhage prognostic
  prediction method using joint-attention interaction mechanism.
\newblock {\em Neural Networks}, 184:107096, 2025.

\bibitem{yu2024scnet}
Xinlei Yu, Ahmed Elazab, Ruiquan Ge, Hui Jin, Xinchen Jiang, Gangyong Jia, Qing
  Wu, Qinglei Shi, and Changmiao Wang.
\newblock Ich-scnet: Intracerebral hemorrhage segmentation and prognosis
  classification network using clip-guided sam mechanism.
\newblock In {\em 2024 IEEE International Conference on Bioinformatics and
  Biomedicine (BIBM)}, pages 2795--2800. IEEE, 2024.

\bibitem{bavcic2024towards}
Boris Ba{\v{c}}i{\'c}, Claudiu Vasile, Chengwei Feng, and Marian~G Ciuc{\u{a}}.
\newblock Towards nation-wide analytical healthcare infrastructures: A
  privacy-preserving augmented knee rehabilitation case study.
\newblock {\em arXiv preprint arXiv:2412.20733}, 2024.

\bibitem{wen2024learning}
Zhibin Wen and Bin Li.
\newblock Learning to unify audio, visual and text for audio-enhanced
  multilingual visual answer localization.
\newblock {\em arXiv preprint arXiv:2411.02851}, 2024.

\bibitem{bi2025prismselfpruningintrinsicselection}
Jinhe Bi, Yifan Wang, Danqi Yan, Xun Xiao, Artur Hecker, Volker Tresp, and
  Yunpu Ma.
\newblock Prism: Self-pruning intrinsic selection method for training-free
  multimodal data selection, 2025.

\bibitem{lu2024robust}
Shilin Lu, Zihan Zhou, Jiayou Lu, Yuanzhi Zhu, and Adams Wai-Kin Kong.
\newblock Robust watermarking using generative priors against image editing:
  From benchmarking to advances.
\newblock {\em arXiv preprint arXiv:2410.18775}, 2024.

\bibitem{li2025set}
Leyang Li, Shilin Lu, Yan Ren, and Adams Wai-Kin Kong.
\newblock Set you straight: Auto-steering denoising trajectories to sidestep
  unwanted concepts.
\newblock {\em arXiv preprint arXiv:2504.12782}, 2025.

\bibitem{xia2022lingyi}
Fei Xia, Bin Li, Yixuan Weng, Shizhu He, Kang Liu, Bin Sun, Shutao Li, and Jun
  Zhao.
\newblock Lingyi: medical conversational question answering system based on
  multi-modal knowledge graphs.
\newblock {\em arXiv preprint arXiv:2204.09220}, 2022.

\bibitem{bi2025llavasteeringvisualinstruction}
Jinhe Bi, Yujun Wang, Haokun Chen, Xun Xiao, Artur Hecker, Volker Tresp, and
  Yunpu Ma.
\newblock Llava steering: Visual instruction tuning with 500x fewer parameters
  through modality linear representation-steering, 2025.

\bibitem{weng2023visual}
Yixuan Weng and Bin Li.
\newblock Visual answer localization with cross-modal mutual knowledge
  transfer.
\newblock In {\em ICASSP 2023-2023 IEEE International Conference on Acoustics,
  Speech and Signal Processing (ICASSP)}, pages 1--5. IEEE, 2023.

\bibitem{zhong2025enhancing}
Jiachen Zhong and Yiting Wang.
\newblock Enhancing thyroid disease prediction using machine learning: A
  comparative study of ensemble models and class balancing techniques.
\newblock 2025.

\bibitem{qiu2025generative}
Shiqing Qiu, Yang Wang, Zong Ke, Qinyan Shen, Zichao Li, Rong Zhang, and
  Kaichen Ouyang.
\newblock A generative adversarial network-based investor sentiment indicator:
  Superior predictability for the stock market.
\newblock {\em Mathematics}, 13(9):1476, 2025.

\bibitem{jing2023multimodal}
Peiguang Jing, Kai Cui, Jing Zhang, Yun Li, and Yuting Su.
\newblock Multimodal high-order relationship inference network for fashion
  compatibility modeling in internet of multimedia things.
\newblock {\em IEEE Internet of Things Journal}, 11(1):353--365, 2024.

\bibitem{ke2025detection}
Zong Ke, Shicheng Zhou, Yining Zhou, Chia~Hong Chang, and Rong Zhang.
\newblock Detection of ai deepfake and fraud in online payments using gan-based
  models.
\newblock {\em arXiv preprint arXiv:2501.07033}, 2025.

\bibitem{he2025enhancing}
Yangfan He, Jianhui Wang, Kun Li, Yijin Wang, Li~Sun, Jun Yin, Miao Zhang, and
  Xueqian Wang.
\newblock Enhancing intent understanding for ambiguous prompts through
  human-machine co-adaptation.
\newblock {\em arXiv preprint arXiv:2501.15167}, 2025.

\bibitem{lu2024mace}
Shilin Lu, Zilan Wang, Leyang Li, Yanzhu Liu, and Adams Wai-Kin Kong.
\newblock Mace: Mass concept erasure in diffusion models.
\newblock In {\em Proceedings of the IEEE/CVF Conference on Computer Vision and
  Pattern Recognition}, pages 6430--6440, 2024.

\bibitem{yi2025score}
Qiang Yi, Yangfan He, Jianhui Wang, Xinyuan Song, Shiyao Qian, Miao Zhang,
  Li~Sun, and Tianyu Shi.
\newblock Score: Story coherence and retrieval enhancement for ai narratives.
\newblock {\em arXiv preprint arXiv:2503.23512}, 2025.

\bibitem{10096391}
Bin Li, Yixuan Weng, Bin Sun, and Shutao Li.
\newblock Learning to locate visual answer in video corpus using question.
\newblock In {\em ICASSP 2023 - 2023 IEEE International Conference on
  Acoustics, Speech and Signal Processing (ICASSP)}, pages 1--5, 2023.

\bibitem{zhang2024improving}
Huan Zhang, Chen Zheng, Yuanjing He, Yan Zhao, and Yuxuan Lai.
\newblock Improving multilingual temporal answering grounding in single video
  via llm-based translation and ocr enhancement.
\newblock In {\em CCF International Conference on Natural Language Processing
  and Chinese Computing}, pages 145--156. Springer, 2024.

\bibitem{zhou2023improving}
Zineng Zhou, Jun Liu, Shuang Cheng, Haiyong Luo, Yang Gu, and Jian Ye.
\newblock Improving cross-modal visual answer localization in chinese medical
  instructional video using language prompts.
\newblock In {\em CCF International Conference on Natural Language Processing
  and Chinese Computing}, pages 221--232. Springer, 2023.

\bibitem{yu2024mqua}
Guyang Yu, Xiaoyang Bi, Jielong Tang, Ming Gu, Tianbai Chen, Zhiqiang Li, and
  Miankuan Zhu.
\newblock Mqua: Multi-level query-video augmentation for multilingual video
  corpus retrieval.
\newblock In {\em CCF International Conference on Natural Language Processing
  and Chinese Computing}, pages 353--364. Springer, 2024.

\bibitem{lei2023two}
Ningjie Lei, Jinxiang Cai, Yixin Qian, Zhilong Zheng, Chao Han, Zhiyue Liu, and
  Qingbao Huang.
\newblock A two-stage chinese medical video retrieval framework with llm.
\newblock In {\em CCF International Conference on Natural Language Processing
  and Chinese Computing}, pages 211--220. Springer, 2023.

\bibitem{ma2024multilingual}
Tianxing Ma, Yueyue Hu, Shuang Jiang, Zhenhao Yin, and Tianning Zang.
\newblock Multilingual temporal answer grounding in video corpus with enhanced
  visual-textual integration.
\newblock In {\em CCF International Conference on Natural Language Processing
  and Chinese Computing}, pages 471--483. Springer, 2024.

\bibitem{cheng2023unified}
Shuang Cheng, Zineng Zhou, Jun Liu, Jian Ye, Haiyong Luo, and Yang Gu.
\newblock A unified framework for optimizing video corpus retrieval and
  temporal answer grounding: fine-grained modality alignment and local-global
  optimization.
\newblock In {\em CCF International Conference on Natural Language Processing
  and Chinese Computing}, pages 199--210. Springer, 2023.

\bibitem{cui2024correlation}
Kai Cui, Shenghao Liu, Wei Feng, Xianjun Deng, Liangbin Gao, Minmin Cheng,
  Hongwei Lu, and Laurence~T Yang.
\newblock Correlation-aware cross-modal attention network for fashion
  compatibility modeling in ugc systems.
\newblock {\em ACM Transactions on Multimedia Computing, Communications and
  Applications}, 2024.

\bibitem{feng2020language}
Fangxiaoyu Feng, Yinfei Yang, Daniel Cer, Naveen Arivazhagan, and Wei Wang.
\newblock Language-agnostic bert sentence embedding.
\newblock {\em arXiv preprint arXiv:2007.01852}, 2020.

\bibitem{li2022vpai_lab}
Bin Li, Yixuan Weng, Fei Xia, Bin Sun, and Shutao Li.
\newblock Vpai\_lab at medvidqa 2022: a two-stage cross-modal fusion method for
  medical instructional video classification.
\newblock In {\em Proceedings of the 21st Workshop on Biomedical Language
  Processing}, pages 212--219, 2022.

\bibitem{jing2023category}
Peiguang Jing, Kai Cui, Weili Guan, Liqiang Nie, and Yuting Su.
\newblock Category-aware multimodal attention network for fashion compatibility
  modeling.
\newblock {\em IEEE Transactions on Multimedia}, 25:9120--9131, 2023.

\end{thebibliography}
\end{document}